# X-ReCoSa: Multi-Scale Context Aggregation for Multi-Turn Dialogue Generation

Danqin Wu

Department of Computer Science, Beijing University of Posts & Telecommunications, Beijing, China

*Abstract*

*In multi-turn dialogue generation, responses are not only related to the topic and background of the context but also related to words and phrases in the sentences of the context. However, currently widely used hierarchical dialog models solely rely on context representations from the utterance-level encoder, ignoring the sentence representations output by the word-level encoder. This inevitably results in a loss of information while decoding and generating. In this paper, we propose a new dialog model X-ReCoSa to tackle this problem which aggregates multi-scale context information for hierarchical dialog models. Specifically, we divide the generation decoder into upper and lower parts, namely the intention part and the generation part. Firstly, the intention part takes context representations as input to generate the intention of the response. Then the generation part generates words depending on sentence representations. Therefore, the hierarchical information has been fused into response generation. we conduct experiments on the English dataset DailyDialog. Experimental results exhibit that our method outperforms baseline models on both automatic metric-based and human-based evaluations.*

## 1. Introduction

Multi-turn dialogue generation task has gained increasing attention in recent years, in which the dialogue contexts usually have complex hierarchical semantic structures. Therefore, it's a challenge to model the complex contexts in a conversation and exploit it to generate coherent and fluent responses [1]. A hierarchical framework has been proposed [2-6], which usually contains a two-level encoder and a response decoder. The two-level encoder sequentially maps the input embedding into sentence representations and context representations. Then the decoder generates responses based on context representations. Many researchers evolve their models under this framework. Hierarchical recurrent encoder-decoder (HRED) [2] adopts Recursive Recurrent Network (RNN) as both dialog encoder and decoder. [3-4] have tried to introduce the attention mechanism into HRED, which uses similarity measures to select relevant information from context. With the self-attention mechanism becoming pervasively used, Transformer [7] is also introduced into dialog models. ReCoSa [5] replaces RNN with Transformer as both utterance-level encoder and response decoder while HSAN [6] takes a further step to use Transformer as a word-level encoder instead of RNN. However, all the aforementioned models decode responses solely depending on context representations, which inevitably results in the loss of sentence-level information.

Context representations contain high-level information such as topic and background while sentence-level semantics have low-level information such as words and phrases. Both of them are important for dialog generation. To make this clear, we give an example in Table 1. Apparently, the topic of responses is related to context1 and context4 while the phrase 'apologize for being late' is associated with 'i'm so late' in context1 and 'we're late' in context2. In this case, solely depending on context representations could lose fine-grained information. And sentence-level semantics could help dialog models generate more related words and phrases.

Table 1. An example from DailyDialog dataset.

| | |
|---|---|
| Context1 | i'm sorry i'm so late .i had a really bad day . |
| Context2 | it's ten after six . we ' re late . but dinner is at six thirty . |
| Context3 | iknow .iknow .i'm really sorry .i lost my bag . |
| Context4 | i'll call the lost and found office. |
| Response | i did n't think of it . thank you . and i do apologize for being late . |

In this paper, we propose a new dialog generation model X-ReCoSa to aggregate both context- and utterance-level semantics into the response generation. Specifically, we split the decoder into two parts, i.e., the intention part and the generation part. We expect that the intention part focuses on the context information to determine the intention of the reply while the generation part needs a fine-grained control over words and phrases to generate responses. the Multi-Head Attention proposed in [7] would be employed to achieve such contexts fusion. The intention part takes context representations as input to multi-head attention and the generating part takes sentence representations as input. With such fusion, we could generate coherent responses from our model. We conduct experiments on the English dataset DailyDialog [8]. The experimental results verify the effectiveness of our method when applied to ReCoSa and HSAN, in terms of automatic metric-based and human judgment.

## 2. RELATED WORK

It's often the case that dialogues in the real world are multi-turn, owing to which multi-turn dialog generation has achieved increasing attention [2,5]. [2] apply a hierarchical RNN-based encoder-decoder model HRED to search prompts and model conversations, which uses a word-level encoder and a context-level encoder to model hierarchical context semantic structures. Since then, many HRED-based variants [9-11] have been widely used in dialogue tasks. But depending solely on the single vector output by RNN would results in semantic loss, generating a large number of generic replies, such as 'I don't know', etc.

To tackle this problem, [3] and [4] try to introduce an attention mechanism, using a specific similarity measure to calculate the attention weight between the reply and the context to select the most relevant context for the current dialogue. However, the above variants are all based on RNN, which has a position bias problem in that RNN is insufficient in modeling long-distant context dependencies and prefers close contexts [5].

With transformers [7] showing powerful capabilities in modeling long-distance dependencies, many researchers begin to apply transformers in multi-turn dialogue models. [5] propose the ReCoSa model, using the transformer as a context-level encoder. HSAN [6] further replaces the word-level encoder with a transformer to build a multi-level self-attention mechanism. Although transformer-based encoders can better dialog contexts, the above models still depend solely on the context representations when decoding. In our paper, we propose a new model called X-ReCoSa to fuse both context- and utterance-level information into response generation, better utilizing hierarchical context information.

## 3. APPROACH

We assume a multi-turn conversation contains $n$ utterances. We take the previous $n-1$ sentences as the dialogue context $\mathbf{c} = \{\mathbf{s}_1, \mathbf{s}_2, \ldots, \mathbf{s}_{n-1}\}$ and the last sentence as a reasonable reply $\mathbf{s}_r = \mathbf{s}_n$ given the contexts. An utterance is defined as $\mathbf{s}_i = \{\mathbf{w}_{i,1}, \mathbf{w}_{i,2}, \ldots, \mathbf{w}_{i,m}\}$, where $\mathbf{w}_{i,m}$ represents the m-th word in the i-th utterance. In this section, we first introduce the architecture of our model X-ReCoSa and then describe the model details.

### 3.1. Architecture

As shown in Figure 1, X-ReCoSa consists of a two-level encoder and response decoder. The word-level encoder adopts Gated Recurrent Unit (GRU) [12] to map each sentence in the context into a high-dimensional representation. And then transformer-based utterance-level encoder projects these sentence representations into context representations. The response decoder is split into two even parts, that is, the intention part and generation part. The two parts take contexts representations and sentence representations as input respectively to generate responses. We will introduce our model in detail in the next section.

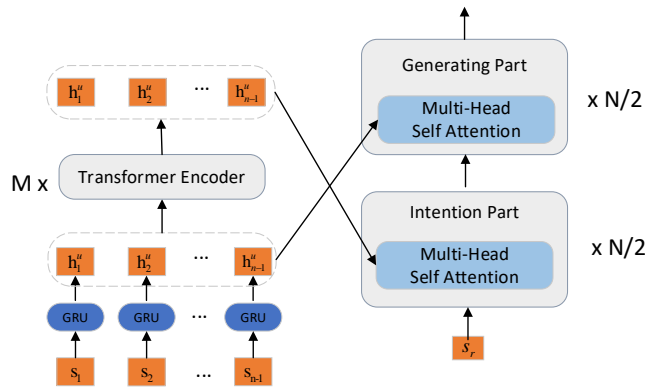

Figure 1. The Architecture of X-ReCoSa.

The left part is the two-level encoder, consisting of GRU-based word-level encoder and Transformer-based utterance-level encoder. The right part are Transformer decoders split into two parts. For simplicity, other modules in Transformer decoder are not visible.

### 3.2. Hierarchical Encoder

We will introduce the hierarchical encoder in detail in this section. According to ReCoSa [5], we adopt a GRU-based word-level encoder and a Transformer-based utterance-level encoder. For each utterance in the given contexts, the word-level encoder adopts the GRU module and inputs a word at each time step $t$, and finally outputs a fixed vector as follows.

$$\mathbf{z}_t = \sigma(W_z \mathbf{x}_t + U_z \mathbf{h}_{t-1}) \qquad (1)$$
$$\mathbf{r}_t = \sigma(W_r \mathbf{x}_t + U_r \mathbf{h}_{t-1}) \qquad (2)$$
$$\overline{\mathbf{h}}_t = tanh(W \mathbf{x}_t + U(\mathbf{r}_t \odot \mathbf{h}_{t-1})) \qquad (3)$$
$$\mathbf{h}_t = (1 - \mathbf{z}_t) \odot \mathbf{h}_{t-1} + \mathbf{z}_t \odot \overline{\mathbf{h}}_t \qquad (4)$$

Where $\mathbf{x}_t$ is embedding of the word $\mathbf{w}_t$, $\mathbf{h}_{t-1}$ is the hidden state at step $t-1$, $\sigma$ represents sigmoid function. $W_z, W_r, W, U_z, U_r, U$ are parameters. We use the final state $\mathbf{h}_m$ as the sentence representation $\mathbf{h}_i^u$. After all utterances in the context pass through GRU, we can obtain sentence representations $\mathbf{H}^u = \{\mathbf{h}_1^u, \mathbf{h}_2^u, \cdots, \mathbf{h}_{n-1}^u\}$.

The Multi-Head Attention in Transformer uses scaled dot-product attention to calculate the relevance score between queries and keys, and output mixed representations.

$$Attention(Q, K, V) = Softmax(\frac{QK^T}{\sqrt{d}})V \qquad (5)$$

Where d is the dimension of hidden representations. we input all the sentence representations as Q, K, V and take the output of the last layer of transformer encoders as context representations $\mathbf{H}^c$.

$$M_i = Attention(\mathbf{H}^u W_i^Q, \mathbf{H}^u W_i^K, \mathbf{H}^u W_i^V) \qquad (6)$$
$$\mathbf{H}^c = Concat(M_1, \ldots, M_H) W^O \qquad (7)$$

Where $M_i \in \mathbb{R}^{n \times d}$, $W^O \in \mathbb{R}^{d \times d}$, $\mathbf{H}^u$ is the sentence representation, H is the number of parallel heads. For each head, we denote learned linear maps by $W_i^Q, W_i^K, W_i^V \in \mathbb{R}^{n \times d/H}$. After all the sentence representations projected, we could get the context representations $\mathbf{H}^c$.

### 3.3. Fusion Decoder

The motivation for our proposed fusion method comes from the observation that people usually first come up with the intention of the response and then organize the words to form the specific content. Therefore, we split the Transformer decoder into two parts, i.e., the intention part and the generation part. The intention part inputs context representations to the Multi-Head Attention, focusing on the higher-level context semantics of the dialog history. The formula is as follows.

$$M_i = Attention(\mathbf{X}_r W_i^Q, \mathbf{H}^c W_i^K, \mathbf{H}^c W_i^V) \qquad (8)$$
$$\mathbf{O}^c = Concat(M_1, \ldots, M_H) W^O \qquad (9)$$

Where $\mathbf{X}_r = \{\mathbf{x}_{r,1}, \mathbf{x}_{r,2}, \ldots, \mathbf{x}_{r,n}\}$. $\mathbf{x}_{r,n}$ means the embedding of the n-th token of the ground truth. We regard $\mathbf{X}_r$ as query Q and $\mathbf{H}^c$ as both the key K and the value V for Multi-Head Attention. The Attention module select proper context semantics to decide the intention of the response to be generated.

The generation part of the decoder focuses on utterance-level information to better organize the words in the generated responses. So, the generation part takes the output from intention part $\mathbf{O}^c$ as query Q and sentence representations $\mathbf{H}^u$ as both key K and value V. Then we could get the final output by following calculation.

$$M_i = Attention(\mathbf{O}^c W_i^Q, \mathbf{H}^u W_i^K, \mathbf{H}^u W_i^V) \qquad (10)$$
$$\mathbf{O}^r = Concat(M_1, \ldots, M_H) W^O \qquad (11)$$

International Journal on Cybernetics & Informatics (IJCI) Vol. 12, No.2, April 2023

With the final output $\mathbf{O}^r$ from generation part, we could map it by a linear module to predict the target response and optimize the loss function as follow.

$$\mathcal{L} = -\log p_\theta(\mathbf{s}_{trg}|\mathbf{c}) \qquad (12)$$

## 4. EXPERIMENTS

### 4.1. Dataset, Baselines and Parameters

Experiments are conducted on a high-quality multi-turn open-domain English dialog dataset DailyDialog. It is manually labeled covering various topics of daily life and contains 13,118 dialogues split into a training set with 11,118 dialogues and validation and test sets with 1000 dialogues each. The dialogue of K utterances can be split into K-1 samples. The last sentence of each example is regarded as the response. Other sentences are taken as dialog history.

We adopt three models as baselines, including HRED, ReCoSa, and HSAN. We compare them with our proposed model X-ReCoSa on both metric-based evaluation and human-based annotation.

For Dailydialog, we take the GPT2 tokenizer for token segmentation, and the vocabulary size is set as 13500. The max length of dialog turns is 10 and the max sentence length is 50. For a fair comparison between all the baselines and our models, the numbers of hidden nodes are all set to 512, batch sizes are set to 32. For optimization, we use AdamW with 1e-3 learning rate.

### 4.2. Evaluation Metrics

To evaluate the quality of the generated replies more comprehensively, we use both automatic evaluation metrics and human judgment following [13]. Automatic metrics can be divided into two kinds. The first kind includes BLEU and Rouge, which evaluate the similarity between generated replies and references based on word overlap. Since BLEU may correlate weakly with human judgments of quality in a single reference setting [14], we employ a multi-reference DailyDialog test set [15] as references to alleviate the problem. The second kind is Distinct-k, which is used to measure the diversity of replies by normalizing the number of distinct k-grams.
For human judgment, given 300 randomly sampled contexts and generated replies, three annotators (all CS majored students) are required to compare X-ReCoSa with baselines and give *win*, *tie* and *lose* labels based on fluency and coherence between contexts and replies. For example, the *win* label for X-ReCoSa means the generated responses are more proper than the baseline models. For fair comparison, Models the responses generated by are blind to all the annotators.

## 5. RESULTS

### 5.1. Metric-based Evaluation

The automatic evaluation results are shown in Table 2. Compared to the original model ReCoSa, our modified model X-ReCoSa achieves significantly higher scores on all the metrics. That means with the fusion of both context and sentence semantics, our model could better utilize dialog history to generate more coherent and diverse responses. Notably, compared with other baseline models, the BLEU scores of our model are significantly higher than that of HRED and HSAN while the Distinct scores of X-ReCoSa are higher than that of HSAN but a little bit lower than HRED, which means our fusion method could improve the coherence between contexts and responses while still maintaining or even improving the diversity of the generated content.

International Journal on Cybernetics & Informatics (IJCI) Vol. 12, No.2, April 2023Table 2. The metric-based evaluation results on DailyDialog with best results bolded (%)

| Model | BLEU-1/2 | BLEU-3/4 | Rouge-1/2 | Rouge-l | Distinct-1/2 |
|---|---|---|---|---|---|
| HRED | 35.01/14.62 | 8.32/5.02 | 4.96/0.24 | 4.93 | **1.08/7.68** |
| ReCoSa | 33.97/13.83 | 8.14/5.25 | 4.75/0.34 | 4.73 | 0.88/4.91 |
| HSAN | 26.66/9.98 | 5.63/3.53 | 0.63/0.01 | 0.63 | 0.09/0.18 |
| X-ReCoSa | **36.7/15.17** | **8.56/5.35** | **8.22/0.88** | **8.16** | 0.98/6.65 |

Table 3. The results of human judgment

| Model | Win(%) | Tie(%) | Lose(%) | Kappa |
|---|---|---|---|---|
| HRED | 36.29 | 37.22 | 26.49 | 0.453 |
| ReCoSa | 35.61 | 43.65 | 20.74 | 0.397 |
| HSAN | 42.96 | 35.13 | 21.91 | 0.412 |

## 5.2. Human Judgment

The results of the human evaluation are shown in Table 3. The percentage of win, loss, and tie are given to evaluate the quality of generated responses by X-ReCoSa. From the results, we can see that the percentage of the *win* label is always larger than that of *lose* label. That shows Our X-ReCoSa outperforms all other models on the DailyDiaog dataset in terms of coherence and fluency. Kappa scores are all greater than 0.375, which means the evaluation results from them could be seen as relatively consistent. After the significance test, we prove that the improvement of our model is significant, i.e., p-value < 0.001.

## 5.3. Case Study

For a better understanding, we give an example in Table 4 to further analyze the quality of generated responses. From the results, we could see that X-ReCoSa generates more related responses than baseline models which only rely on context representations. In the example, baseline models are likely to generate common responses like 'what's your problem?' and 'i'm sorry to hear that', etc., but our model could contain phrases related to contexts like 'got a fever'. The reason is that our model considers not only high-level context semantics but also low-level sentence information. The multi-scale contexts are fused to give rich features for the response decoder to generate words.

Table 4. The generated responses from different models.

| | |
|---|---|
| Context1 | good morning . what's the matter with you ? |
| Context2 | good morning , doctor . i have a terrible headache . |
| Ground Truth | all right , young man . tell me how it got started . |
| HRED | what's your problem ? |
| ReCoSa | i'm sorry to hear that . |
| HSAN | i'm sorry , but i'm not sure . |
| X-ReCoSa | have you got a fever ? |

## 6. CONCLUSIONS

In this paper, we propose a new model X-ReCoSa to exploit both context and utterance representations using multi-head attention. Experimental results show that X-ReCoSa could improve the coherence between responses and dialog contexts. We could conclude that our fusion method could better utilize hierarchical dialog contexts and benefit the model performance.

## REFERENCES

[1] Kashif Khan, Gaurav Sahu, Vikash Balasubramanian, Lili Mou & Olga Vechtomova, (2020)". Adversarial Learning on the Latent Space for Diverse Dialog Generation", In Proceedings of the 28th International Conference on Computational Linguistics, pp 5026–5034.

[2] Iulian V. Serban, Alessandro Sordoni, Yoshua Bengio, Aaron Courville & Joelle Pineau, (2016) "Building end-to-end dialogue systems using generative hierarchical neural network models", In Proceedings of the Thirtieth AAAI Conference on Artificial Intelligence, pp. 3776–3783.

[3] Zhiliang Tian, Rui Yan, Lili Mou, Yiping Song, Yansong Feng & Dongyan Zhao, (2017) "How to Make Context More Useful? An Empirical Study on Context-Aware Neural Conversational Models", In Proceedings of the 55th Annual Meeting of the Association for Computational Linguistics, Vol. 2, pp. 231–236.

[4] Chen Xing, Yu Wu, Wei Wu, Yalou Huang & Ming Zhou, (2018) "Hierarchical recurrent attention network for response generation", In Proceedings of the Thirty-Second AAAI Conference on Artificial Intelligence and Thirtieth Innovative Applications of Artificial Intelligence Conference and Eighth AAAI Symposium on Educational Advances in Artificial Intelligence, pp. 5610–5617

[5] Hainan Zhang, Yanyan Lan, Liang Pang, Jiafeng Guo & Xueqi Cheng, (2019) "ReCoSa: Detecting the Relevant Contexts with Self-Attention for Multi-turn Dialogue Generation", In Proceedings of the 57th Annual Meeting of the Association for Computational Linguistics, pp. 3721–3730.

[6] Yawei Kong, Lu Zhang, Can Ma & Cong Cao, (2021) "HSAN: A Hierarchical Self-Attention Network for Multi-Turn Dialogue Generation", 2021 IEEE International Conference on Acoustics, Speech and Signal Processing (ICASSP), pp. 7433-7437.

[7] Ashish Vaswani, Noam Shazeer, Niki Parmar, Jakob Uszkoreit, Llion Jones, Aidan N. Gomez, Łukasz Kaiser & Illia Polosukhin, (2017) "Attention is all you need", In Proceedings of the 31st International Conference on Neural Information Processing Systems (NIPS'17), pp. 6000–6010.

[8] Yanran Li, Hui Su, Xiaoyu Shen, Wenjie Li, Ziqiang Cao & Shuzi Niu, (2017) "DailyDialog: A Manually Labelled Multi-turn Dialogue Dataset", In Proceedings of the Eighth International Joint Conference on Natural Language Processing, Vol. 1, pp. 986–995.

[9] Iulian V. Serban, Alessandro Sordoni, Ryan Lowe, Laurent Charlin, Joelle Pineau, Aaron Courville & Yoshua Bengio, (2017) "A hierarchical latent variable encoder-decoder model for generating dialogues", In Proceedings of the Thirty-First AAAI Conference on Artificial Intelligence (AAAI'17), pp.3295–3301.

[10] Yookoon Park, Jaemin Cho & Gunhee Kim, (2018) "A Hierarchical Latent Structure for Variational Conversation Modeling", In Proceedings of the 2018 Conference of the North American Chapter of the Association for Computational Linguistics: Human Language Technologies, Vol. 1, pp. 1792–1801.

[11] Lei Shen, Yang Feng & Haolan Zhan, (2019) "Modeling Semantic Relationship in Multi-turn Conversations with Hierarchical Latent Variables", In Proceedings of the 57th Annual Meeting of the Association for Computational Linguistics, pp. 5497–5502.

[12] Kyunghyun Cho, Bart van Merriënboer, Caglar Gulcehre, DzmitryBahdanau, Fethi Bougares, Holger Schwenk & Yoshua Bengio, (2014) "Learning Phrase Representations using RNN Encoder–Decoder for Statistical Machine Translation", In Proceedings of the 2014 Conference on Empirical Methods in Natural Language Processing (EMNLP), pp. 1724–1734.

[13] Jiamin Wang, Xiao Sun, Qian Chen & Meng Wang, (2022) "Information-Enhanced Hierarchical Self-Attention Network for Multiturn Dialog Generation", in IEEE Transactions on Computational Social Systems, pp. 1-12.

[14] Chia-Wei Liu, Ryan Lowe, Iulian Serban, Mike Noseworthy, Laurent Charlin & Joelle Pineau, (2016) "How NOT To Evaluate Your Dialogue System: An Empirical Study of Unsupervised Evaluation Metrics for Dialogue Response Generation", In Proceedings of the 2016 Conference on Empirical Methods in Natural Language Processing, pp. 2122–2132.

[15] Prakhar Gupta, Shikib Mehri, Tiancheng Zhao, Amy Pavel, Maxine Eskenazi & Jeffrey Bigham, (2019) "Investigating Evaluation of Open-Domain Dialogue Systems With Human Generated Multiple References", In Proceedings of the 20th Annual SIGdial Meeting on Discourse and Dialogue, pp. 379–391.

## AUTHOR

**Danqin Wu** was born in 1998 and received the bachelor's degree from Central University of Finances and Economics, Beijing, China, in 2020. He is pursing the master's degree with the school of Computer, Beijing University of Posts and Telecommunications, Beijing, China. His main research filed is dialogue generation

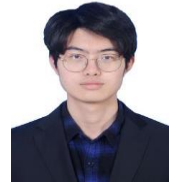